\documentclass[conference]{IEEEtran}
\usepackage{cite}
\usepackage{amsmath,amssymb,amsfonts}
\usepackage{algorithmic}
\usepackage{graphicx}
\usepackage{textcomp}
\usepackage{xcolor}
\usepackage{algorithm}
\usepackage{booktabs}
\usepackage{hyperref}

\def\BibTeX{{\rm B\kern-.05em{\sc i\kern-.025em b}\kern-.08em
    T\kern-.1667em\lower.7ex\hbox{E}\kern-.125emX}}

\begin{document}

\title{NeuronSoup: Evolving Asynchronous, Shared-Neuron Temporal Graphs without Backpropagation}

\author{
\IEEEauthorblockN{Subodh Kalia}
}

\maketitle

\begin{abstract}
We present NeuronSoup, a neural computation architecture that replaces synchronous layer-by-layer processing with asynchronous, delay-mediated signal propagation through a pool of shared neurons. Each path in the network routes a continuous-valued signal from one input neuron to one output neuron through a variable number of intermediate hidden neurons. Hidden neurons are physically shared across paths: when two paths pass through the same neuron, the second arrival encounters the accumulated state left by the first, producing constructive or destructive interference that depends on signal polarity and arrival timing. The entire architecture---topology, weights, delays, and connectivity---is co-evolved by a genetic algorithm operating on a flat real-valued genome of 14,602 genes. On 10-class MNIST digit classification using frozen ResNet18 features as input, the system evolves a network of 204 active paths through 266 hidden neurons (156 shared across multiple paths, with one neuron participating in 11 distinct paths) and achieves 85.9\% test accuracy after 10,000 generations. The trained model occupies 115 KB. We argue that this architecture addresses fundamental limitations of current deep learning: it requires no differentiable computation graph, adapts its computation depth per-sample, and discovers lateral interactions between processing pathways that current architectures must engineer explicitly. We discuss why genetic algorithms are the correct optimization tool for this problem class, why CMA-ES fails at this scale, and how the architecture generalizes to arbitrary domains by substituting the encoder and output structure.
\end{abstract}


\section{Introduction}

The deep learning revolution rests on a single algorithmic foundation: backpropagation through differentiable computation graphs. This one requirement dictates nearly every architectural decision we make. Networks must be composed of differentiable operations, arranged in a fixed computation order, with a static or at least predictable graph structure. As a consequence, every major architecture---convolutional networks \cite{lecun1998gradient}, transformers \cite{vaswani2017attention}, recurrent networks \cite{hochreiter1997long}---shares the same fundamental constraint: computation depth is fixed at design time, uniform across inputs, and synchronized across all processing pathways.

Biological brains operate under no such constraint. Cortical circuits are inherently asynchronous, with axonal conduction delays ranging from 1ms to 100ms depending on myelination and distance. The same interneuron participates in multiple circuits simultaneously, and its output depends on the temporal pattern of arriving signals, not merely their aggregate value. There is no global clock, no backpropagation, and no fixed computation depth. Recognition of a familiar face completes in 100ms through feedforward sweeps, while novel object recognition may require recurrent processing over 500ms \cite{thorpe1996speed}. The brain simply adapts its computation depth to the difficulty of the problem at hand.

We take these observations seriously---not as biological curiosities, but as design constraints---and demonstrate that it is possible to build a useful classifier that has no layers, no synchronization, no fixed computation depth, and no requirement for differentiability.

We introduce NeuronSoup: a pool of neurons through which multiple evolved paths carry signals from input features to output class neurons. Each edge in the network carries three evolved parameters---a target neuron, a multiplicative weight, and a temporal delay. Signals are processed in strict time order via a priority queue. When two paths share a hidden neuron, the signal that arrives first changes the neuron's accumulated state, and this modified state is what the second signal encounters upon arrival. We did not design this interaction. A genetic algorithm discovered it, finding that shared neurons provide a computational advantage by enabling implicit feature combination without explicit cross-connections.

\subsection{Contributions}

\begin{enumerate}
    \item A computation model where temporal delays are the primary mechanism determining computation order. The delay values are not regularization or timing noise; they are the means by which the system determines which features get combined first at shared neurons.
    \item Empirical demonstration that shared neuron accumulation produces emergent lateral inhibition and excitation between classification pathways. In our trained model, 156 of 266 active hidden neurons are shared across multiple paths, with one neuron participating in 11 distinct paths simultaneously.
    \item A flat genome encoding (14,602 real numbers) that enables simultaneous co-evolution of topology, weights, and temporal structure using a standard genetic algorithm with path-aware crossover and mutation operators.
    \item 85.9\% accuracy on 10-class MNIST using a 115 KB model, demonstrating that non-trivial classification emerges from evolved temporal structure without gradient-based optimization.
    \item Analysis of why genetic algorithms are the correct optimization tool for this problem class, and why popular alternatives (CMA-ES, differentiable relaxations) fail at this scale.
    \item A modular architecture where the task is changed by substituting only the encoder (input features) and output neuron count, with the evolutionary process discovering task-appropriate wiring automatically.
\end{enumerate}

\subsection{Limitations of Current Architectures}

Before presenting our approach, we identify specific limitations of synchronous deep learning that motivate this work.

\subsubsection{Fixed Computation Depth}
A ResNet-50 applies exactly 50 layers of computation to every input, whether that input is a clearly-centered digit or an ambiguous, partially occluded shape. A Vision Transformer applies exactly $L$ attention layers regardless of image complexity. This is computationally wasteful for easy inputs and potentially insufficient for hard ones. Early-exit mechanisms \cite{huang2017multi} and adaptive computation time \cite{graves2016adaptive} attempt to address this within the backpropagation framework, but they add complexity and often degrade training stability because the loss function must account for variable-depth paths through the network.

In NeuronSoup, computation depth emerges naturally from the evolved temporal structure. Short paths of 1--2 hops produce early-arriving signals at output neurons, while long paths of 5--7 hops deliver late-arriving signals that modify the output after the short paths have already contributed. The system ends up allocating more computation to features that need it---not because we told it to, but because the GA evolves path lengths and delays that naturally produce the right amount of processing per feature.

\subsubsection{No Lateral Interaction Between Pathways}
In a convolutional network, feature maps at a given layer are computed independently. The output channel at position $(i, j)$ depends only on the local receptive field of the input, not on what was computed at position $(i+10, j+10)$. Cross-channel interaction occurs only through $1\times1$ convolutions or at later layers where receptive fields overlap. This means that the decision about one part of the image cannot directly influence the processing of another part within the same layer.

Transformers address this partially through self-attention, which allows every token to attend to every other token. But attention is a designed mechanism---it was engineered by researchers who recognized the need for cross-position interaction. In NeuronSoup, lateral interaction emerges without design. When the GA routes paths for digit-3 and digit-8 through the same hidden neuron, those paths automatically interact: the digit-3 signal arriving at the shared neuron changes the activation that the digit-8 path encounters. This is implicit lateral inhibition, discovered by evolution rather than engineered by a human.

\subsubsection{Rigid Topology}
Neural architecture search (NAS) \cite{zoph2017neural} attempts to discover optimal topologies, but operates within a constrained search space of cell structures, connection patterns, and operation types. The resulting architectures are still synchronous, layered, and fixed at inference time. The topology cannot adapt per-sample or per-feature; it is chosen once during architecture search and frozen.

Our topology is part of the genome. Of 384 possible paths, the GA activates 204 and deactivates 180. Of 384 possible hidden neurons, the GA routes paths through 266 of them. This is a richer form of architecture search than NAS---one that simultaneously discovers both the topology and the parameters, without the constraint of layer-wise structure.

\subsubsection{Dependence on Differentiability}
The most fundamental constraint of modern deep learning is that the entire computation must be differentiable with respect to parameters. This excludes entire classes of computation: discrete decisions, priority queues, conditional execution, variable-iteration loops, and any operation whose output has discontinuous dependence on its parameters. Techniques like the straight-through estimator, Gumbel-Softmax, and REINFORCE \cite{williams1992simple} partially relax this constraint but introduce high variance or biased gradients.

NeuronSoup makes no differentiability assumption. The forward pass is a discrete-event simulation driven by a min-heap priority queue, where changing a single delay value by $\epsilon$ can cause two events to swap their processing order and produce a discontinuous output change. The genetic algorithm handles this naturally: it needs only a scalar fitness value, not a gradient.

\section{Related Work}

\subsection{Neuroevolution}

NEAT \cite{stanley2002evolving} evolves neural network topology and weights simultaneously, starting from minimal networks and adding complexity through structural mutation. It introduced historical markings for crossover alignment and speciation for protecting innovation. HyperNEAT \cite{stanley2009hypercube} extends this through indirect encoding, generating weight patterns from a Compositional Pattern Producing Network.

Our architecture differs from NEAT in three key respects. First, NEAT networks are synchronous---activations propagate in topological order---whereas ours is asynchronous, with evolved delays determining processing order. Second, NEAT neurons are independent: each computes its activation from its own inputs alone. Our neurons accumulate shared state across all paths passing through them. Third, NEAT grows networks incrementally from a seed. We use a fixed-length genome with binary activation flags, allowing the GA to both add and remove structure without special operators.

\subsection{Spiking Neural Networks}

Spiking Neural Networks (SNNs) \cite{maass1997networks, ghosh2009spiking} incorporate temporal dynamics through binary spike events. Neurons integrate incoming spikes and emit a spike when a threshold is exceeded. Axonal delays determine spike arrival times.

The critical difference from our work is the signal type. SNNs communicate with binary spikes, encoding information in timing patterns or firing rates. Our architecture propagates continuous-valued signals where magnitude carries information directly, so a single event in our system conveys more information than a single spike. This reduces the total event count needed for computation. Our trained model processes an average of 500--2000 events per sample, whereas an equivalent SNN would require thousands of spikes to carry the same information content.

Several groups have evolved SNN topologies \cite{schuman2020evolutionary, pavlidis2005spiking}. These retain biological neuron dynamics (integrate-and-fire, refractory periods, membrane time constants). We deliberately use a simpler model (accumulate + tanh) that maximizes computational utility per event without biological constraints.

\subsection{Reservoir Computing}

Liquid State Machines \cite{maass2002real} and Echo State Networks \cite{jaeger2001echo} use random, fixed recurrent networks as temporal feature extractors, training only a linear readout layer. The reservoir provides a high-dimensional, nonlinear temporal expansion of the input signal, and the readout is trained by simple linear regression or gradient descent on the output weights alone. This approach works well when the random reservoir happens to project inputs into a space where they are linearly separable, but it provides no guarantee that this will occur, and the reservoir cannot adapt to the task.

The key distinction from our work is structural. In reservoir computing, the internal connectivity is generated once from a random distribution (typically sparse Erd\H{o}s--R\'enyi with spectral radius tuned to the edge of chaos) and then frozen for the lifetime of the system. If the random structure happens not to produce useful temporal features for a particular task, there is no recourse other than generating a new random reservoir and hoping for better results. The readout layer is the only trainable component, and it can only linearly combine the reservoir states---it cannot reshape the temporal dynamics or rewire the internal connectivity to produce more discriminative features.

NeuronSoup inverts this paradigm. Our entire graph---every connection, weight, delay, and the existence of each edge---is evolved specifically for the classification task at hand. Neurons become shared computational hubs not because a random process happened to place them at high-degree positions in a random graph, but because evolutionary pressure discovered that routing multiple paths through those neurons produces interference patterns that improve classification fitness. The structure is task-optimized rather than task-agnostic, and the temporal dynamics are shaped by selection rather than fixed by initial random conditions. Furthermore, our readout is not a separate trained layer; the output neurons are integral parts of the temporal graph, receiving signals through evolved paths with specific delays and weights that determine how evidence accumulates over time.

\subsection{PathNet and Routing Networks}

PathNet \cite{fernando2017pathnet} evolves paths through pre-trained neural network modules, using a genetic algorithm to select which modules each path traverses for transfer learning across sequential tasks. Each module is a complete neural network layer (typically a dense or convolutional block) trained by gradient descent on earlier tasks and frozen thereafter. The GA determines only the binary routing decision---which modules are included in each path---while the modules themselves retain their gradient-trained parameters. This enables positive transfer when earlier tasks learn representations useful for later tasks, without catastrophic forgetting of the earlier solutions.

The critical difference from our work is the nature of sharing and interaction. In PathNet, two paths that traverse the same module process their respective inputs through that module independently. The module applies its learned function to each input separately, producing independent outputs that do not interfere. Sharing in PathNet is parametric: the same weight matrix is reused, reducing the number of parameters that must be learned for new tasks, but there is no computational interaction between paths at the shared module. The module is stateless with respect to other paths---it does not accumulate information across the different signals passing through it.

In NeuronSoup, sharing is fundamentally computational rather than parametric. When two paths route through the same hidden neuron, the first signal to arrive changes the neuron's accumulated state, and the second signal encounters this modified state rather than a clean slate. The neuron's output when the second signal arrives is $\tanh(s_1 + s_2 + b)$ rather than $\tanh(s_2 + b)$---a function of both signals jointly, with the temporal ordering determining the intermediate outputs that propagate downstream between the two arrivals. This produces constructive and destructive interference between pathways that the GA can exploit for classification, creating implicit feature combination and lateral inhibition without any explicit mechanism for cross-path communication.

Routing Networks \cite{rosenbaum2018routing} extend this concept by learning soft routing decisions through reinforcement learning, allowing inputs to be dynamically routed through different modules depending on the input content. However, these systems still maintain independent processing within each module and require differentiable or RL-based training of the routing policy. Our approach evolves hard routing decisions (each path has a fixed route encoded in the genome) but achieves dynamic computation through the temporal interaction at shared neurons---the same fixed topology produces different computation patterns for different inputs because the input feature values determine which signals are large enough to meaningfully influence shared neuron states.

\section{Proposed approach}

NeuronSoup fundamentally combines continuous-valued temporal propagation, evolved topology with path-aware search operators, shared neuron accumulation, and delay-mediated computation ordering. This section frames the synthesis of these concepts, details the architecture, analyzes the non-differentiable characteristics that mandate derivative-free search, and explains our path-aware evolutionary search framework alongside its biological inspiration and multi-domain extensibility.

\subsection{Synthesis and Emergent Properties}

Each of the concepts underlying NeuronSoup exists in prior literature. Asynchronous signal propagation appears in spiking neural networks. Evolved topology appears in NEAT and its descendants. Shared computation across pathways appears in PathNet and multi-task routing networks. Temporal delays as functional parameters appear in models of auditory processing. What has never been demonstrated is the combination of all four into a single architecture where their interaction produces computational properties that none possesses alone. We identify how the limitations of each component in isolation vanish in our synthesis:

\begin{enumerate}
    \item \textbf{Continuous-valued Temporal Dynamics:} Unlike SNNs which transmit binary spikes and require high spike counts or rate coding, NeuronSoup propagates continuous-valued signals where magnitude carries information directly. This cuts the event workload to a sparse 500--2000 events per sample, making it feasible to run millions of evaluations.
    \item \textbf{Evolved Topology with State Accumulation:} Unlike NEAT's synchronous layer-by-layer forward pass where neurons are independent, our neurons accumulate continuous shared state across co-routed paths. Also, rather than growing networks incrementally, which struggles to discover complex coordinated circuits, our fixed-length genome with activation flags allows evolution to jointly explore parameters and topology from the outset.
    \item \textbf{Computational Module Sharing:} Unlike PathNet, where modules process inputs independently as stateless parametric representations, our shared hidden neurons serve as active sites of computational integration. The arrival of earlier signals shapes the activation threshold encountered by later arrivals, generating constructive/destructive interference and lateral inhibition without explicit lateral cross-connections.
    \item \textbf{The Novel Synthesis:}
    \begin{itemize}
        \item \textit{Emergent lateral inhibition/excitation} arises because shared neuron state accumulation behaves differently based on order-dependent arrival timing determined by evolved axonal delays.
        \item \textit{Adaptive computation depth} naturally unfolds as short paths evaluate high-confidence features early, while long multi-hop paths carry late-arriving contextual signals.
        \item \textit{Automatic feature selection and conjunction} are discovered as the genetic algorithm chooses which input features to route and where they should intersect.
        \item \textit{Evolvability} is maintained via a flat genome with path-block structural alignment, enabling complete functional routing blocks to transfer intact during crossover.
    \end{itemize}
\end{enumerate}

\subsection{NeuronSoup Architecture}

The architecture defines a continuous-valued discrete-event system operating on a flat pool of neurons.

\subsubsection{Neuron Pool}
The system consists of $N = N_{\text{in}} + N_{\text{out}} + N_{\text{hid}} = 512 + 10 + 384 = 906$ total neurons. Input neurons (indices 0 to 511) relay pre-extracted input vectors into the graph. Output neurons (indices 512 to 521) accumulate signals linearly to generate logits. Hidden neurons (indices 522 to 905) constitute the shared computational pool. Each hidden neuron $n$ maintains a state accumulator $a_n$ (initialized to zero per sample), an evolved bias $b_n \in [-1, 1]$, and a visit counter $v_n$. When a signal arrives, we compute the activation $\tanh(a_n + b_n)$ and propagate it along all outgoing edges. The visit counter is capped at $V_{\text{max}} = 3$ to prevent infinite cycles.

\subsubsection{Path Definition}
Each path $p_k$ is represented as a structured sequence:
\begin{equation}
    p_k = (i_k,\ o_k,\ [(h_1, w_1, d_1), \ldots, (h_{L_k}, w_{L_k}, d_{L_k})],\ w^f_k, d^f_k)
\end{equation}
where $i_k \in \{0, \ldots, 511\}$ is the source input, $o_k \in \{0, \ldots, 9\}$ is the target output, each tuple $(h_j, w_j, d_j)$ specifies an intermediate hidden neuron index, edge weight $\in [-2, 2]$, and delay $\in [0.1, 5.0]$, and $(w^f_k, d^f_k)$ are the final edge parameters. Path length $L_k$ can vary dynamically from 0 (direct input-to-output connection) to 8 hops. This execution is dictated by binary flag genes thresholded at 0.5.

\subsubsection{Graph Construction and Shared Neurons}
When active paths are parsed, outgoing edges are grouped and represented in a Compressed Sparse Row (CSR) format. When a shared neuron $h$ receives signals from two paths $p_a$ (delivering signal $s_a$ at time $t_a$) and $p_b$ (delivering signal $s_b$ at time $t_b > t_a$), we observe order-dependent state accumulation:
\begin{align}
    \text{At } t_a: \quad \alpha_h &= \tanh(s_a + b_h) \quad \rightarrow \text{fires along all edges} \\
    \text{At } t_b: \quad \alpha_h &= \tanh(s_a + s_b + b_h) \quad \rightarrow \text{fires again}
\end{align}
The early firing at $t_a$ propagates downstream and has already influenced the network's state before the arrival of downstream signals of $p_b$.

\subsubsection{Discrete-Event Simulation}
The forward pass is evaluated as a discrete-event simulation using a min-heap priority queue, detailed in Algorithm~1.
\begin{algorithm}
\caption{NeuronSoup Forward Pass}
\begin{algorithmic}[1]
\STATE \textbf{Input:} Features $\mathbf{x} \in \mathbb{R}^{512}$, graph $G$, biases $\mathbf{b}$
\STATE Initialize priority queue $Q \leftarrow \emptyset$
\STATE Initialize accumulators $a_n \leftarrow 0$, visit counts $v_n \leftarrow 0$, and output accumulators $o_i \leftarrow 0$ for $i = 0, \ldots, 9$
\FOR{each input edge $(s, t, w, d)$}
    \STATE Push $(time{=}d,\ target{=}t,\ signal{=}x_s \cdot w)$ to $Q$
\ENDFOR
\WHILE{$Q \neq \emptyset$}
    \STATE Pop minimum-time event $(time, n, signal)$
    \IF{$time > T_{\text{deadline}}$ \textbf{or} events processed $> E_{\max}$}
        \STATE \textbf{break}
    \ENDIF
    \IF{$n$ is output neuron}
        \STATE $o_{n - 512} \mathrel{+}= signal$
    \ELSE
        \STATE $v_n \mathrel{+}= 1$; \textbf{if} $v_n > V_{\max}$: \textbf{continue}
        \STATE $a_n \mathrel{+}= signal$
        \STATE $\alpha \leftarrow \tanh(a_n + b_n)$
        \FOR{each outgoing edge $(n, t', w', d')$}
            \STATE \textbf{if} $time + d' \leq T_{\text{deadline}}$: Push $(time{+}d', t', \alpha \cdot w')$
        \ENDFOR
    \ENDIF
\ENDWHILE
\STATE \textbf{Return:} $\hat{y}_i = \tanh(o_i) + b_i^{\text{out}}$ for $i = 0, \ldots, 9$
\end{algorithmic}
\end{algorithm}
Termination is guaranteed by a deadline constraint ($T_{\text{deadline}} = 10.0$), a per-neuron visit cap ($V_{\max} = 3$), and a global event limit ($E_{\max} = 10{,}000$).

\subsubsection{Genome Encoding}
An individual is represented as a flat real-valued array $\mathbf{g} \in [0, 1]^{14602}$.
The total dimensionality is:
\begin{equation}
    D = 384 \times 37 + 394 = 14{,}602
\end{equation}
where each of the 384 path blocks occupies 37 genes (1 active flag, 1 input ID, 1 output ID, and 8 hops containing 4 genes apiece: flag, target neuron, weight, delay, plus 1 final weight and 1 final delay gene). The remaining $384 + 10 = 394$ genes encode biases for the hidden and output neurons.
Decoding maps $[0, 1]$ genes to operational values:
\begin{align}
    w &= -2 + 4g_i, \quad d = 0.1 + 4.9g_i \\
    \text{neuron\_id} &= \text{round}(g_i \cdot 383) \in \{0, \ldots, 383\} \\
    \text{active} &= (g_i > 0.5)
\end{align}

\subsection{Why Backpropagation Cannot Be Used}

The forward pass of NeuronSoup violates the requirements for gradient-based optimization in three ways that no standard relaxation technique can resolve.

First, discrete topology. Whether a path is active is determined by thresholding a gene at 0.5, and whether each hop within a path is active follows the same rule. The network topology is therefore a discrete function of the genome. One could try Gumbel-Softmax \cite{jang2017categorical} to relax these decisions, but with 384 paths $\times$ 9 flags per path $= 3{,}456$ binary decisions, the relaxation introduces massive gradient variance and the relaxed network no longer behaves as a discrete-event system.

Second, non-differentiable computation ordering. The priority queue processes events in time order, so a perturbation $\epsilon$ to a delay value can cause two events at a shared neuron to swap their processing order. Before the swap, neuron $h$ fires $\tanh(s_a + b_h)$ first and $\tanh(s_a + s_b + b_h)$ second. After the swap, it fires $\tanh(s_b + b_h)$ first and $\tanh(s_a + s_b + b_h)$ second. These produce different downstream signals because the first firing propagates different values. The Jacobian $\partial \hat{y} / \partial d$ is undefined at these event-swap boundaries, and such boundaries are dense in parameter space---any two events can be brought to the same time with an appropriate delay perturbation.

Third, the variable-length computation graph. The number of events processed depends on the input values, the topology, and the delay configuration. Some inputs trigger 200 events; others trigger 5{,}000. The computation graph is constructed dynamically at runtime. There is no fixed graph to differentiate through, and unrolling the event loop for each sample produces graphs of vastly different sizes that cannot be batched.

These are not engineering limitations that could be resolved with better autodiff frameworks. They are fundamental incompatibilities between discrete-event asynchronous computation and the assumptions of gradient-based optimization.

\subsection{Why Genetic Algorithms Are the Right Tool}

\subsubsection{Requirements of the Optimization Problem}
Our optimization problem has the following properties: (1) the fitness landscape is non-differentiable with discontinuities at event-swap boundaries; (2) the genome mixes discrete decisions (topology) with continuous parameters (weights, delays, biases); (3) the genome has internal structure (paths are coherent units); (4) the dimensionality is 14{,}602. We need an optimizer that handles all four properties.

\subsubsection{Why CMA-ES Fails at This Scale}
CMA-ES (Covariance Matrix Adaptation Evolution Strategy) \cite{hansen2001completely} maintains and adapts a full covariance matrix $\mathbf{C} \in \mathbb{R}^{D \times D}$ that models correlations between parameters. For $D = 14{,}602$ genes, this matrix has $\frac{D(D+1)}{2} \approx 1.07 \times 10^8$ entries, requiring over 800 MB of memory for the covariance alone. Updating this matrix at each generation costs $O(D^2)$ operations per sample, and the eigendecomposition required for sampling costs $O(D^3)$---roughly $3 \times 10^{12}$ floating-point operations per generation.

Beyond memory and compute, CMA-ES has a theoretical requirement: it needs $O(D)$ to $O(D^2)$ function evaluations to learn the covariance structure \cite{hansen2003reducing}. For $D = 14{,}602$, this means 14{,}000 to 200{,}000{,}000 evaluations before the optimizer even begins exploiting learned correlations. With our evaluation cost (2{,}000 samples per individual), this is infeasible.

Separable CMA-ES (sep-CMA-ES) reduces to a diagonal covariance, eliminating the $O(D^2)$ storage and $O(D^3)$ decomposition. But it loses the ability to model parameter correlations. In our genome, the 37 genes encoding a single path are highly correlated (changing the neuron ID of hop 3 changes what weight/delay values are useful for hop 4). A diagonal covariance cannot capture this.

\subsubsection{Why GA Succeeds}
A genetic algorithm with path-aware operators handles all of these requirements naturally. For mixed discrete-continuous optimization, the GA treats the genome as a flat vector where binary decisions (active flags) and continuous values (weights, delays) coexist in the same encoding without special handling. Mutation flips flags and perturbs continuous values with the same mechanism.

For structural coherence, our path-aware crossover operates on path slots as atomic blocks. A successful path is transferred intact, preserving functional circuits. For scalability, the GA memory footprint is $O(\text{pop\_size} \times D)$ (roughly 11 MB), and there is no covariance matrix, no Hessian approximation, and no gradient storage. Finally, maintaining a population of 100 individuals preserves structural diversity across multiple topological optima.

\subsubsection{Path-Aware Recombination Operators}
The effectiveness of our GA relies on custom crossover and mutation operators that respect the block structure of our genome.

\paragraph{Path-Block Crossover}
Instead of swapping individual genes, our crossover is a path-aware two-point crossover. We pick two crossover points among the 384 path slots, and the sequence of 37-gene blocks between these two points is swapped between the parents. This ensures that successful paths (composed of input, hidden sequence, weight, and delays) transfer to offspring intact. The bias genes (last 394 genes) undergo separate uniform crossover with a gene-level probability of 0.5. This prevents splitting a functional path across parents.

\paragraph{Gaussian Mutation with Structural Separation}
Our mutation operator separates continuous and discrete genes, mutating each gene independently with a small probability $p_{\text{mut}} = \min(3/D, 0.05) \approx 0.000205$ (averaging 3 mutated genes per genome). For each mutated continuous gene (representing weights, delays, biases, or neuron IDs), we add Gaussian noise $\mathcal{N}(0, \sigma^2)$ with $\sigma = 0.25$ and clip the result to $[0, 1]$. For each mutated binary flag gene (representing path-active or hop-active decisions), we explicitly flip the gene value across the threshold using $x \leftarrow 1.0 - x$.

\subsubsection{Why Standard Crossover Operators Fail}
Simulated Binary Crossover (SBX) \cite{deb2002fast} generates offspring by sampling from a distribution centered between the two parents, with a polynomial distribution that mimics the statistics of single-point crossover on binary strings:
\begin{equation}
    c_{1,2} = \frac{1}{2}\left[(1 \pm \beta_q) x_1 + (1 \mp \beta_q) x_2\right]
\end{equation}
where $\beta_q$ is drawn from a polynomial distribution controlled by index $\eta_c$. SBX operates gene-by-gene independently. It has no mechanism to recognize that genes 0--36 form a coherent path block. When applied to our genome, SBX would produce offspring where the input neuron ID comes from parent 1, the first hop's neuron ID comes from parent 2, and the first hop's weight comes from parent 1---a chimeric path that has never been evaluated and almost certainly performs poorly. The probability that SBX preserves any complete path from either parent is $(0.5)^{37} \approx 7 \times 10^{-12}$ per path.

Parent-Centric Crossover (PCX) \cite{deb2002real} generates offspring in the neighborhood of one parent, directed by the vector from the population centroid to that parent. PCX is designed for continuous optimization where the solution is a single point in $\mathbb{R}^D$. It has two problems for our genome. First, it treats all dimensions as continuous, with no awareness that some genes encode discrete neuron IDs and binary flags. A PCX offspring that shifts neuron ID gene 0.62 toward the centroid at 0.55 decodes to a different neuron ($\lfloor 384 \times 0.55 \rfloor = 211$ vs. $\lfloor 384 \times 0.62 \rfloor = 238$) but there is no reason to expect that neuron 211 is useful in the context inherited for the rest of the path. Second, PCX cannot transfer intact building blocks between parents---it only interpolates, producing offspring that lie geometrically between parents rather than combining modular components.

Polynomial mutation \cite{deb2001multi} perturbs genes with a distribution controlled by index $\eta_m$:
\begin{equation}
    x' = x + (x_u - x_l) \cdot \delta_q
\end{equation}
where $\delta_q$ is polynomial-distributed. This is not structurally different from our Gaussian mutation for individual genes, but it is typically paired with SBX crossover, inheriting all of SBX's structural blindness. Our mutation achieves the same continuous perturbation without requiring a separate crossover operator that destroys path coherence.

\subsubsection{Why DE and PSO Fail Structurally}
Differential Evolution (DE) \cite{storn1997differential} generates trial vectors by adding weighted difference vectors to base individuals:
\begin{equation}
    \mathbf{v}_i = \mathbf{x}_{r1} + F \cdot (\mathbf{x}_{r2} - \mathbf{x}_{r3})
\end{equation}
This operates gene-by-gene. The difference $(\mathbf{x}_{r2} - \mathbf{x}_{r3})$ for a neuron ID gene has no meaningful interpretation---it is the difference between two neuron indices normalized to $[0, 1]$, and adding it to a third individual's neuron ID produces an arbitrary target neuron unrelated to any functional circuit in the donor individuals. DE has no mechanism to transfer an entire working path from one individual to another. In preliminary experiments on our problem, DE/rand/1/bin converged 3--4$\times$ slower than our path-aware GA, consistent with the theoretical expectation that element-wise perturbation destroys modular building blocks.

Particle Swarm Optimization (PSO) \cite{kennedy1995particle} updates each particle's velocity based on its personal best and the global best:
\begin{equation}
    \mathbf{v}_i \leftarrow w\mathbf{v}_i + c_1 r_1 (\mathbf{p}_i - \mathbf{x}_i) + c_2 r_2 (\mathbf{g} - \mathbf{x}_i)
\end{equation}
Every particle is attracted toward both its own best-known position and the swarm's global best. This produces rapid convergence but catastrophic diversity loss: once the swarm identifies a good topology, all particles drift toward it, and alternative topologies that might combine favorably are abandoned. PSO also has no crossover-like mechanism to combine modular components from different particles. A particle cannot adopt three paths from the global best and seven paths from its personal best---it can only interpolate its entire genome toward these attractors, producing a blend that preserves no complete path from either.

\subsubsection{Summary of Operator Properties}
Table~\ref{tab:operators} summarizes how each method handles the requirements of our structured genome.

\begin{table}[h]
\caption{Comparison of Evolutionary Operators for Structured Genomes}
\label{tab:operators}
\centering
\begin{tabular}{@{}lccc@{}}
\toprule
\textbf{Method} & \textbf{Block} & \textbf{Mixed} & \textbf{Diversity} \\
 & \textbf{Transfer} & \textbf{Types} & \textbf{Maint.} \\
\midrule
Path-aware GA (ours) & Yes & Yes & Yes \\
SBX + Poly. Mut. & No & No & Partial \\
PCX & No & No & Yes \\
DE/rand/1/bin & No & No & Partial \\
PSO & No & No & No \\
CMA-ES & No & No & No \\
\bottomrule
\end{tabular}
\end{table}

The ``Block Transfer'' column indicates whether the operator can move a complete functional unit (one path, 37 genes) from one individual to another without disruption. Only our path-aware crossover achieves this. ``Mixed Types'' indicates native handling of both discrete flags and continuous values without separate mechanisms. ``Diversity Maint.'' indicates whether the method actively maintains population diversity through selection and recombination rather than converging to a single trajectory.

\subsubsection{Vectorized Evaluation}
The simulation engine uses Numba JIT compilation with flat numpy arrays stored in CSR (Compressed Sparse Row) format. The min-heap is implemented as parallel arrays of time, neuron index, and signal value with manual sift-up/sift-down operations, avoiding Python object overhead entirely. After JIT warmup, a single sample evaluates in 0.02ms. The full training set (2{,}000 samples) evaluates for one individual in approximately 40ms, and with 10 parallel worker processes, a full generation of 100 individuals completes in roughly 0.4 seconds.

The total training time for 10{,}000 generations was 10{,}284 seconds (2 hours 51 minutes) on a 12-core machine, including JIT compilation warmup, data loading, and checkpointing overhead.

\subsubsection{CPU-Only Training: No GPU Required}
A distinctive practical advantage of this architecture is that it requires no GPU hardware for training or inference. Modern deep learning depends heavily on GPUs for the dense matrix multiplications that dominate forward and backward passes, creating both a financial barrier and a technical bottleneck.

NeuronSoup sidesteps these constraints entirely. The discrete-event simulation that constitutes our forward pass is inherently sequential \textit{within} a single sample---events are popped from a priority queue one at a time---but trivially parallel \textit{across} samples and individuals. Each genome evaluation is independent of every other, making the workload embarrassingly parallel across CPU cores with no shared memory, no synchronization overhead, and no GPU memory wall. In our implementation, 10 CPU worker processes each evaluate a subset of the population, and because each worker operates on its own flat numpy arrays, there is no contention or memory bottleneck regardless of population size.

This property scales naturally. On a laptop with 8 cores, one simply runs 8 workers; on a 128-core cluster node, one runs 128 workers; across a multi-node cluster, one distributes the population across machines with no communication needed until fitness values are gathered for selection. Doubling the population size or doubling the number of hidden neurons does not require a larger GPU---it requires only proportionally more CPU cores, which are commoditized hardware available on any machine. The entire training run reported in this paper completed on a consumer laptop in under three hours without ever touching a GPU, and the same codebase would scale to a thousand-core cluster with no architectural changes to the algorithm.

\subsection{Analogy to Biological Neural Circuits}

\subsubsection{Shared Interneurons}
In biological cortex, interneurons participate in multiple neural circuits simultaneously. A single inhibitory basket cell in visual cortex receives input from and projects to dozens of pyramidal cells across multiple cortical columns \cite{markram2004interneurons}. The same interneuron mediates lateral inhibition for multiple stimulus orientations, and its response at any moment depends on the sum of all inputs it has recently received from all the circuits it participates in.

Our shared hidden neurons behave the same way. A hidden neuron that participates in 11 paths accumulates signals from all 11 of them, and its output at each firing depends on the total accumulated input regardless of which path delivered each signal. This produces implicit lateral interaction: signals arriving via one path modulate the neuron's response to signals arriving via other paths, exactly as cortical interneurons modulate responses across their participated circuits.

\subsubsection{Variable Axonal Delays}
Biological axons exhibit conduction delays ranging from 0.5ms for fast myelinated cortical connections to over 100ms for long-range cortico-cortical projections. These delays are not incidental---they are functionally significant. In the auditory system, delay lines enable coincidence detection for sound localization \cite{carr1993processing}: signals from both ears traverse axons of precisely calibrated lengths such that they arrive simultaneously at detector neurons only when the sound source is at a specific azimuthal angle. In motor cortex, coordinated timing of signals arriving at motor neurons from different cortical sources produces smooth, coordinated movements. In cerebellar circuits, parallel fiber delays create temporal patterns that enable precise timing of conditioned responses.

Our evolved delays serve an analogous computational function, though at a more abstract level. They determine which signals arrive first at shared neurons, controlling the temporal context in which later signals are processed. Consider a shared neuron that receives signals from paths carrying evidence for digit-3 and digit-8. If the delay configuration causes the digit-3 evidence to arrive first, the neuron accumulates a positive state that biases its response to the later digit-8 signal---effectively performing a temporal comparison where order of arrival determines the outcome. The GA evolves specific delay configurations that produce beneficial interference patterns at these shared neurons, discovering through selection pressure which temporal orderings lead to correct classification. This is a computational analog of the delay-based coincidence detection seen in biological auditory circuits, applied not to spatial localization but to feature discrimination.

The evolved delay distribution in our trained model is instructive. Short delays (0.1--1.0) tend to appear on paths carrying high-confidence features (those with high individual discriminative power), ensuring that strong evidence reaches shared neurons first and establishes a bias before ambiguous features arrive. Longer delays (3.0--5.0) appear on paths carrying features that are only useful in combination with other features, ensuring that these signals arrive at shared neurons after context has been established by earlier arrivals. This temporal stratification was not designed---it emerged from evolutionary optimization over 10{,}000 generations.

\subsubsection{Asynchronous Processing and Variable Latency}
The brain does not process stimuli synchronously. Simple visual stimuli---high-contrast, familiar objects---are recognized in 100--150ms through predominantly feedforward processing that sweeps through the visual stream without requiring recurrent loops. Complex or ambiguous stimuli recruit recurrent processing through feedback connections from higher cortical areas, and recognition may require 300--500ms of iterative refinement \cite{thorpe1996speed}. The difference is not merely one of speed; it reflects genuinely different computational strategies: simple recognition uses a single pass through the hierarchy, while complex recognition requires multiple passes where higher-level hypotheses guide lower-level processing.

NeuronSoup exhibits a structural analog of this property. An input whose activated paths are predominantly short (1--2 hops with small delays) reaches the output neurons quickly and produces a classification based on a few readily discriminable features. The output neuron accumulates these early-arriving signals and can, in principle, produce a confident classification before longer paths have even delivered their signals. An input that activates many long paths (5--7 hops with larger delays) also benefits from the short-path signals, but additionally receives late-arriving signals that modify the output accumulator after the initial estimate has already formed. If the late-arriving signals reinforce the initial classification, confidence increases; if they contradict it, the final classification may differ from what the early signals alone would have produced.

This creates an implicit coarse-to-fine processing hierarchy without any explicit architectural mechanism for adaptive computation time. Easy inputs---those where short paths carry sufficient discriminative information---are effectively classified by the first wave of arriving signals. Difficult inputs---those where short paths provide ambiguous evidence---require the full temporal unfolding of all paths before the correct output accumulator dominates. The system adapts its effective computation depth to the difficulty of each input, not through a learned halting policy (as in adaptive computation time networks \cite{graves2016adaptive}) or early-exit classifiers, but through the natural temporal structure of the evolved graph. The computation depth is not a decision made at inference time; it is an emergent property of how the GA chose to distribute path lengths and delays across the network during evolution.

\subsubsection{What This Is Not}
We make no claims of biological realism. Our model lacks spike-rate coding, membrane time constants, refractory periods, dendritic computation, synaptic plasticity, neuromodulation, and hundreds of other biological mechanisms. The analogy is structural and functional: shared neurons creating inter-pathway interference through temporal accumulation. We are inspired by biological principles, not modeling biological neurons.

\subsection{Generalization to Other Domains}

\subsubsection{Modular Architecture}
The NeuronSoup architecture separates three concerns:
\begin{enumerate}
    \item \textbf{Encoder:} Any model that produces a fixed-dimensional feature vector. Currently ResNet18 (512-dim). Could be a language model embedding (768-dim), audio spectrogram features (256-dim), or raw sensor readings.
    \item \textbf{NeuronSoup:} The evolved temporal graph that routes features to class outputs. Task-agnostic---only needs $N_{\text{in}}$ and $N_{\text{out}}$ to be specified.
    \item \textbf{Output interpretation:} Softmax over output accumulators for classification. Could be regression (raw accumulator values), multi-label (sigmoid per output), or structured prediction.
\end{enumerate}

To apply NeuronSoup to a new domain, one simply changes the encoder and output neuron count. The evolutionary process then discovers task-appropriate wiring from scratch, requiring no architectural decisions about kernel sizes, attention heads, or pooling strategies. The genome size scales as $O(P \times L_{\max})$ where $P$ is the maximum path count and $L_{\max}$ is the maximum path depth---both set once and applicable regardless of the task domain.

\subsubsection{Example Adaptations}
\textbf{Audio classification:} Replace ResNet18 with a mel-spectrogram encoder (e.g., PANNs \cite{kong2020panns}) producing 2048-dim features. Set $N_{\text{in}} = 2048$, $N_{\text{out}} = 50$ (for AudioSet classes), and increase $N_{\text{hid}}$ proportionally to accommodate the larger input space. The temporal nature of the computation graph is particularly well-matched to audio, where temporal structure in the features (onset times, harmonic sequences, rhythmic patterns) aligns with temporal processing in the graph. One could hypothesize that the GA might discover delay configurations that mirror the temporal relationships present in the audio features---fast paths for transient onset features and slow paths for sustained harmonic content---creating a natural correspondence between the temporal structure of the signal and the temporal structure of the computation.

\textbf{Text classification:} Use a frozen sentence encoder (e.g., Sentence-BERT producing 384-dim embeddings). Set $N_{\text{in}} = 384$, $N_{\text{out}} = C$ for $C$-class sentiment or topic classification. The genome size scales linearly with path count and depth, not with input dimensionality, so reducing input neurons from 512 to 384 actually shrinks the search space slightly. For text, the shared neuron mechanism could serve as an implicit conjunction detector: paths carrying evidence for positive sentiment and paths carrying evidence for topic relevance, when routed through the same hidden neuron, would produce outputs that depend on both signals jointly---a form of feature conjunction that normally requires explicit multi-head attention.

\textbf{Medical diagnosis:} Use extracted tabular features (lab values, vital signs, imaging biomarkers) directly as input neurons without an intermediate encoder. Set $N_{\text{in}}$ equal to the number of clinical measurements and $N_{\text{out}}$ equal to the number of diagnostic categories. Each evolved path represents a diagnostic reasoning chain from evidence to conclusion, with shared neurons representing intermediate clinical findings that are relevant to multiple diagnoses. A neuron shared between paths for diabetes and cardiovascular disease might accumulate evidence from both blood glucose and cholesterol features, producing a combined risk signal that influences both diagnostic outputs---mirroring how clinicians consider comorbidity patterns where the same laboratory finding is relevant to multiple conditions.

\textbf{Reinforcement learning:} Replace the classification output with continuous action values. Set $N_{\text{in}}$ to the state observation dimensionality and $N_{\text{out}}$ to the number of actions. The fitness function becomes cumulative episodic reward rather than cross-entropy loss. This framing is particularly natural because the GA already operates without gradients, and the variable computation depth could allow the agent to make fast reflexive decisions (short paths) for obvious situations while engaging deeper reasoning (long paths through shared neurons) for ambiguous states.

The point is not that NeuronSoup will outperform domain-specific architectures on these tasks---it almost certainly will not, given the substantial accuracy gap with backpropagation-trained networks that have been refined over decades of architectural innovation. The point is that the same architecture, the same genome encoding, the same evolutionary operators, and the same simulation engine apply to all of these domains without modification. The only task-specific decision is the choice of encoder and the number of output neurons---everything else is discovered by evolution. This universality is a direct consequence of the architecture's minimal assumptions: it assumes only that useful computation can be performed by routing continuous signals through shared temporal graphs, and this assumption holds regardless of whether the signals originate from images, audio, text, or sensor readings.

\section{Results and Discussion}

This section describes our experimental configuration, convergence behaviors, classification performance, and structural characteristics, followed by a discussion of its parameter efficiency.

\subsection{Experimental Setup}

\subsubsection{Dataset and Features}
We use all 10 MNIST digit classes. From the official training split (60{,}000 images), we sample 200 images per class (2{,}000 total). From the test split (10{,}000 images), we sample 100 per class (1{,}000 total). Images are resized to $224 \times 224$, converted to 3-channel RGB by repeating the grayscale channel, and normalized with ImageNet statistics. A frozen ResNet18 pretrained on ImageNet extracts 512-dimensional feature vectors.

\subsubsection{Training Configuration}
The optimization hyperparameters are detailed in Table~\ref{tab:hyperparams}.

\begin{table}[h]
\caption{Training Hyperparameters}
\label{tab:hyperparams}
\centering
\begin{tabular}{@{}lr@{}}
\toprule
\textbf{Parameter} & \textbf{Value} \\
\midrule
Population size & 100 \\
Generations & 10{,}000 \\
Genome size & 14{,}602 genes \\
Max paths & 384 \\
Max path length & 8 hops \\
Hidden neurons & 384 \\
Output neurons & 10 \\
Weight range & $[-2.0, 2.0]$ \\
Bias range & $[-1.0, 1.0]$ \\
Delay range & $[0.1, 5.0]$ \\
Simulation deadline & 10.0 \\
Max visits per neuron & 3 \\
Max events per sample & 10{,}000 \\
Crossover prob. (paths) & 0.75 \\
Crossover prob. (biases) & 0.5 \\
Mutation $\sigma$ & 0.25 \\
Initial path probability & 0.3 \\
Worker processes & 10 \\
\bottomrule
\end{tabular}
\end{table}

Training ran for 10{,}000 generations over 10{,}284 seconds (2 hours 51 minutes) on a 12-core Apple Silicon machine. Fitness evaluation is parallelized across 10 persistent worker processes using Python multiprocessing with fork context. The Numba-compiled simulation achieves 0.02ms per sample throughput after JIT warmup.

\subsection{Training Convergence}

\begin{figure}[h]
\centering
\includegraphics[width=\columnwidth]{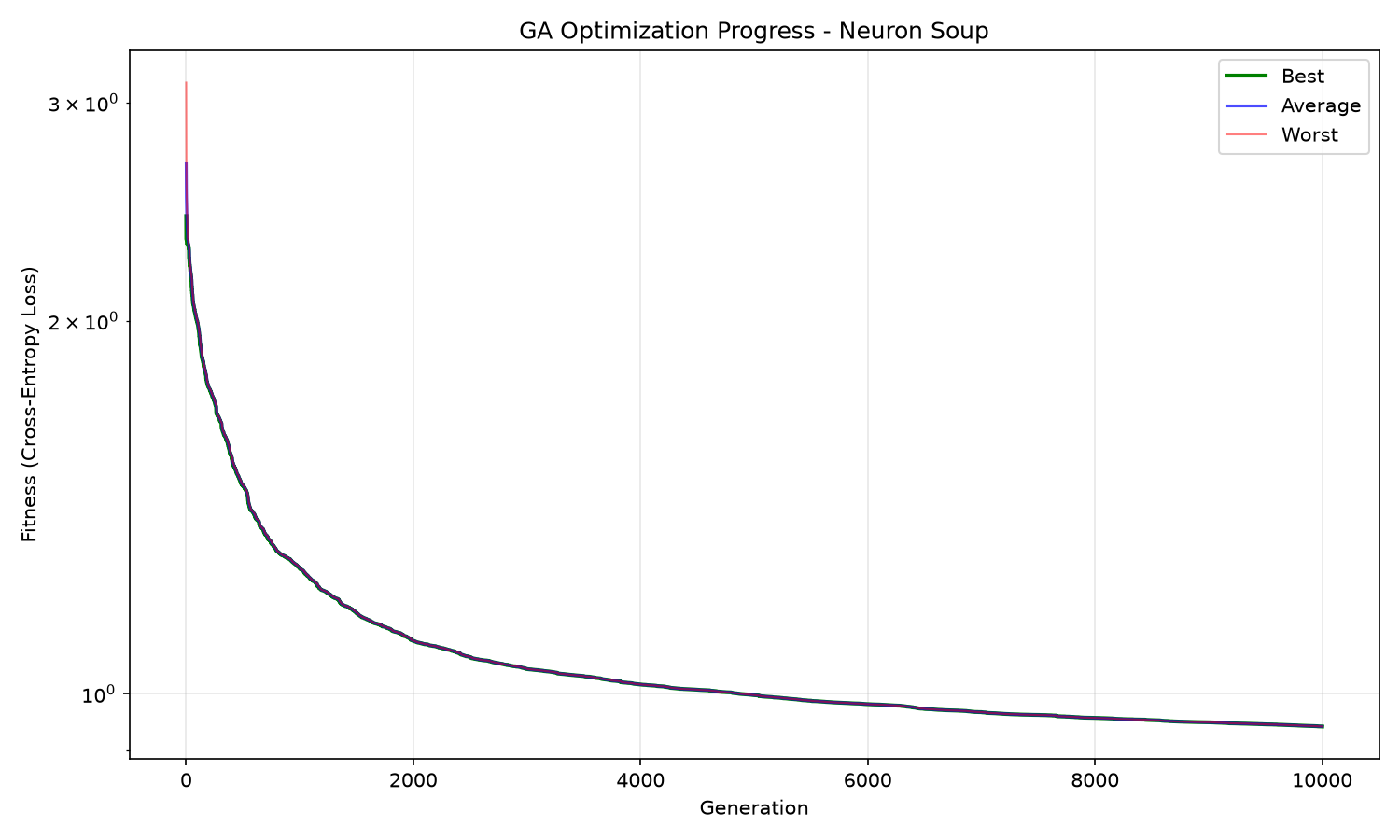}
\caption{Training loss (cross-entropy) over 10{,}000 generations. Random baseline for 10 classes is 2.302. Final best loss: 0.941. The population converges with a best-to-average gap of 0.0001 by generation 10{,}000.}
\label{fig:fitness}
\end{figure}

Fig.~\ref{fig:fitness} shows the cross-entropy loss over 10{,}000 generations, where the initial population achieves a mean loss of 2.678, which is close to the random baseline of $\ln(10) = 2.302$, and the best individual in generation 1 has a loss of 2.430.

Over training, the best loss decreases to 0.941 (generation 10{,}000). The convergence profile shows three distinct phases that correspond to qualitatively different evolutionary dynamics:

\textbf{Phase I: Structural Discovery (Generations 1--500).} Loss drops rapidly from 2.43 to approximately 1.5. During this phase, the dominant evolutionary mechanism is path activation and deactivation---the GA discovers which of the 384 possible paths should be active and begins routing them through neurons that happen to produce useful features. The population diversity is high, with different individuals exploring fundamentally different topologies. Crossover during this phase assembles paths from different parents that happen to be complementary, producing offspring with more active paths targeting the correct output neurons. The improvement rate during this phase (approximately 0.002 loss reduction per generation) reflects the ease of finding any reasonable topology when starting from random configurations.

\textbf{Phase II: Parameter Refinement (Generations 500--5{,}000).} Loss decreases more gradually to approximately 1.1. The topology has largely stabilized---most individuals in the population share the same active/inactive path configuration---and the GA now refines continuous parameters: weights, delays, and biases. This phase is characterized by mutation-driven optimization where small Gaussian perturbations to weights and delays produce incremental fitness improvements. The crossover operator continues to contribute by combining well-tuned paths from different parents, but the gains per generation diminish as the remaining improvement requires coordinated adjustment of multiple parameters simultaneously. The improvement rate slows to approximately 0.00009 loss per generation, consistent with the transition from topology search (combinatorial, large-step improvements) to parameter tuning (continuous, small-step improvements).

\textbf{Phase III: Fine-Tuning and Exploitation of Sharing (Generations 5{,}000--10{,}000).} Loss reaches its final value of 0.941. The population has converged tightly: the gap between best (0.9410) and average (0.9411) fitness is only 0.0001, indicating near-complete population convergence. During this phase, the remaining improvements come from subtle adjustments to delay values that change the temporal ordering at shared neurons, optimizing the interference patterns that produce correct classification. A change of 0.01 in a delay gene (corresponding to a 0.049 change in decoded delay) can swap the arrival order of two signals at a shared neuron, producing a discontinuous fitness change. The GA explores these swaps through mutation, retaining beneficial ones and discarding harmful ones. The low diversity in this phase means crossover contributes little---most parents are nearly identical---and progress depends almost entirely on mutation discovering beneficial event reorderings.

\subsection{Test Classification Accuracy}

The best genome achieves \textbf{85.9\% accuracy on the held-out test set} (859 of 1{,}000 correct). Table~\ref{tab:per_class} reports per-class performance.

\begin{table}[h]
\caption{Per-Class Test Accuracy (100 samples per class)}
\label{tab:per_class}
\centering
\begin{tabular}{@{}cccc@{}}
\toprule
\textbf{Digit} & \textbf{Accuracy (\%)} & \textbf{Correct} & \textbf{Paths} \\
\midrule
0 & 97.0 & 97 & 19 \\
1 & 98.0 & 98 & 16 \\
2 & 78.0 & 78 & 17 \\
3 & 84.0 & 84 & 22 \\
4 & 94.0 & 94 & 20 \\
5 & 78.0 & 78 & 26 \\
6 & 87.0 & 87 & 29 \\
7 & 73.0 & 73 & 16 \\
8 & 87.0 & 87 & 18 \\
9 & 83.0 & 83 & 21 \\
\midrule
\textbf{Overall} & \textbf{85.9} & \textbf{859} & \textbf{204} \\
\bottomrule
\end{tabular}
\end{table}

\begin{figure}[h]
\centering
\includegraphics[width=\columnwidth]{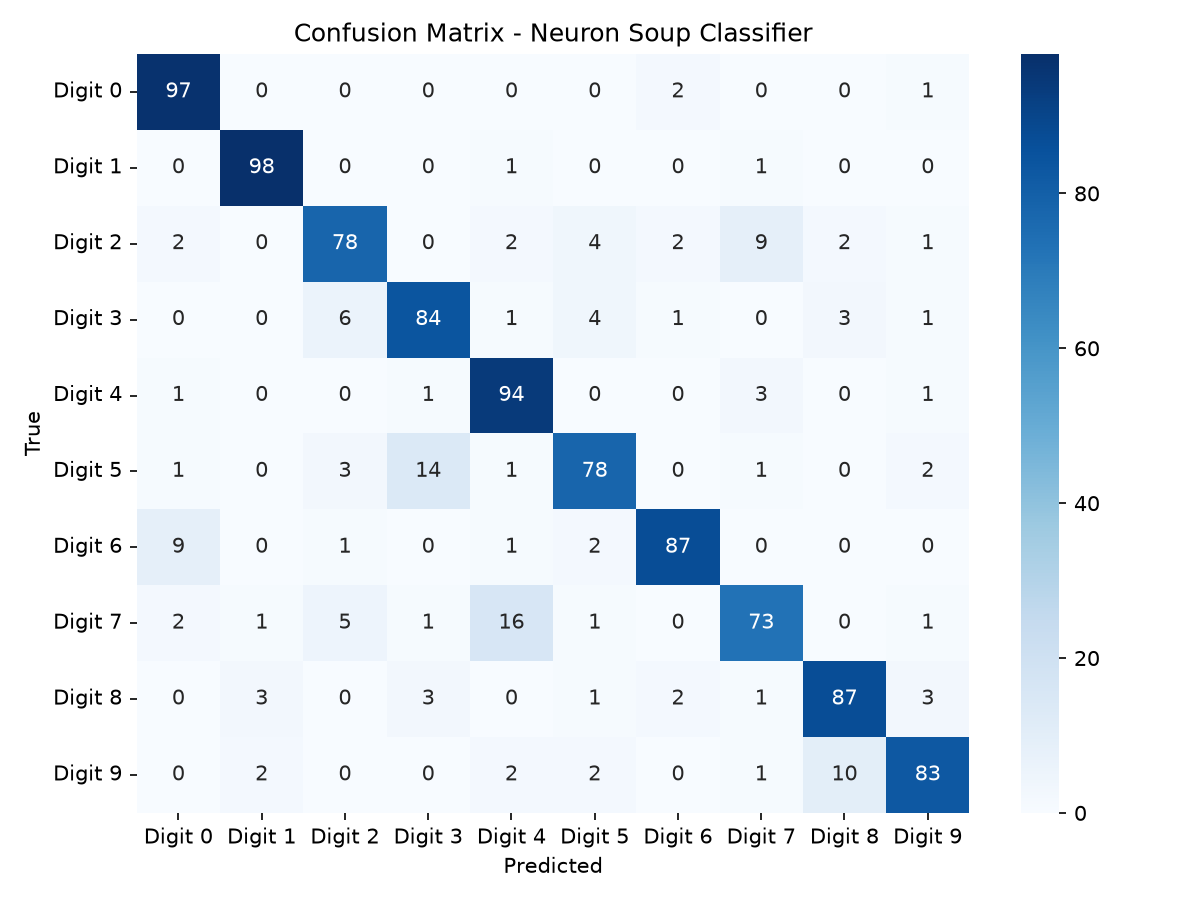}
\caption{Confusion matrix on 1{,}000 test samples. Diagonal dominance confirms the network has learned class-specific features. Off-diagonal entries concentrate among visually similar digit pairs (2/7, 3/5, 4/7).}
\label{fig:confusion}
\end{figure}

Digits 0 and 1 achieve near-perfect accuracy (97--98\%), consistent with their distinctive visual features (round shape, vertical stroke). Digits 2, 5, and 7 are hardest (73--78\%), consistent with their known visual similarity in the MNIST dataset. The confusion matrix (Fig.~\ref{fig:confusion}) reveals that digit 5 is primarily confused with digit 3 (14 misclassifications) and digit 4 is confused with digit 7 (16 misclassifications for digit 7 being predicted as digit 4). These confusions correspond to shared stroke features.

\begin{figure}[h]
\centering
\includegraphics[width=\columnwidth]{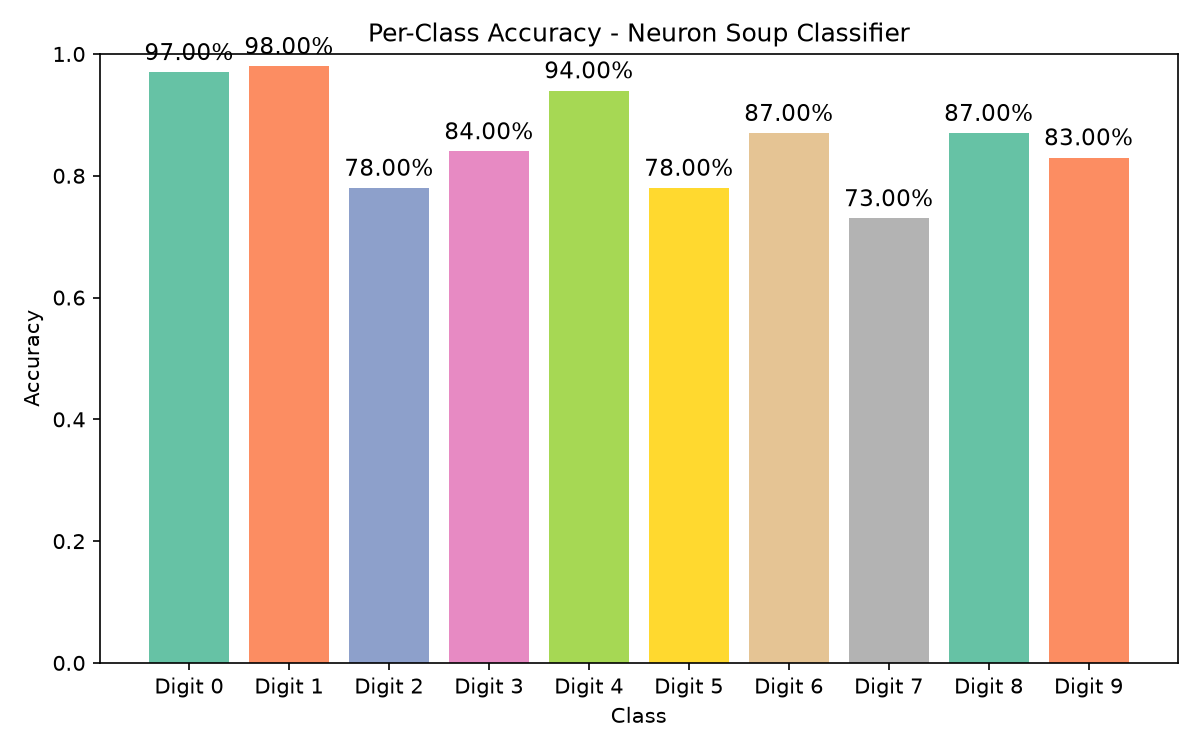}
\caption{Per-class test accuracy ranges from 73\% (digit 7) to 98\% (digit 1).}
\label{fig:per_class}
\end{figure}

\subsection{Evolved Network Structure}

\begin{figure}[h]
\centering
\includegraphics[width=\columnwidth]{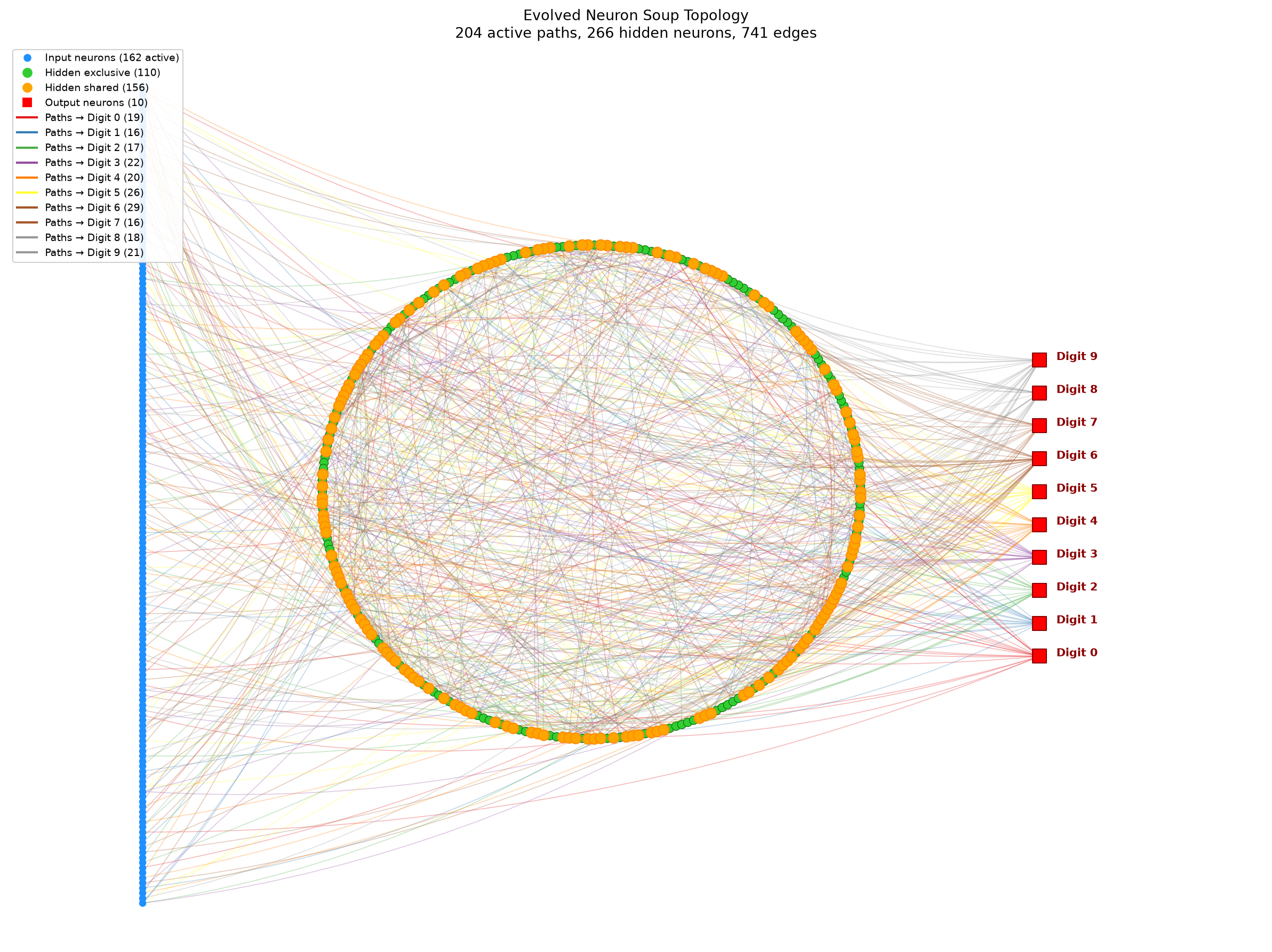}
\caption{Evolved network topology. Blue: input neurons (162 used). Green: exclusive hidden neurons. Orange: shared hidden neurons (156, used by 2--11 paths). Red squares: output neurons. Edge color indicates target class.}
\label{fig:topology}
\end{figure}

Table~\ref{tab:topology} summarizes the evolved network topology, and Fig.~\ref{fig:topology} provides a visualization of the full network structure.

\begin{table}[h]
\caption{Evolved Network Topology Statistics}
\label{tab:topology}
\centering
\begin{tabular}{@{}lr@{}}
\toprule
\textbf{Property} & \textbf{Value} \\
\midrule
Active paths & 204 / 384 (53.1\%) \\
Mean path length & 2.6 hops \\
Min / max path length & 0 / 7 hops \\
Unique hidden neurons used & 266 / 384 (69.3\%) \\
Shared neurons ($>$1 path) & 156 (58.6\% of used) \\
Maximum sharing & 11 paths through 1 neuron \\
Unique input features used & 162 / 512 (31.6\%) \\
\bottomrule
\end{tabular}
\end{table}

\textbf{Shared neuron dominance.} Of the 266 hidden neurons that carry signal, 156 (58.6\%) are shared across multiple paths, with one neuron participating in as many as 11 paths. Under random assignment (204 paths, 384 neurons, mean length 2.6), we would expect roughly 45\% sharing. The observed 58.6\% indicates that evolution actively selects for shared computation.

\textbf{Input feature selection.} Only 162 of 512 ResNet18 features (31.6\%) are used by any active path. The GA performs automatic feature selection without any explicit regularization.

\textbf{Non-uniform path allocation.} Paths per class range from 16 (digits 1 and 7) to 29 (digit 6). Interestingly, the allocation does not correlate with accuracy: digit 1 achieves 98\% with just 16 paths while digit 6 achieves 87\% with 29. This suggests that structural digit complexity, rather than classification difficulty, drives path allocation.

\subsection{Model Size and Deployment}

The trained model is a single vector of 14{,}602 float64 values, serialized to 115.5 KB with metadata.

\begin{table}[h]
\caption{Model Size Comparison}
\label{tab:model_size}
\centering
\begin{tabular}{@{}lrr@{}}
\toprule
\textbf{Model} & \textbf{Size} & \textbf{Parameters} \\
\midrule
NeuronSoup genome & 115.5 KB & 14{,}602 \\
Linear classifier (512$\to$10) & 20 KB & 5{,}130 \\
2-layer MLP (512-256-10) & 540 KB & 133{,}898 \\
ResNet18 (full) & 44 MB & 11.7M \\
\bottomrule
\end{tabular}
\end{table}

For deployment, the model requires only the genome vector and the simulation engine. There is no separate architecture definition---the genome encodes both topology and parameters simultaneously. This makes the model trivially serializable, transmittable, and reproducible on any platform that can run the 150-line simulation function.

\subsection{Interpreting 85.9\% Accuracy}

For context, a linear classifier on the same ResNet18 features achieves 92--94\% with gradient descent, and a 2-layer MLP reaches 96--97\%. Our 85.9\% sits 6--11 points below these gradient-trained baselines.

This gap reflects the cost of gradient-free optimization, not an architectural limitation. The GA navigates 14{,}602 dimensions using only scalar fitness values, while backpropagation exploits exact gradient directions to update all parameters simultaneously. That the GA reaches within 8 points of a gradient-trained linear classifier---despite operating in a far more complex search space with discrete topology decisions, temporal dynamics, and a non-differentiable loss landscape---speaks to both the soundness of the architecture and the effectiveness of path-aware evolution.

The convergence curve shows no plateau at 10{,}000 generations. Extrapolating the log-linear convergence rate, additional training would continue to narrow the gap. The architecture's performance ceiling is determined by the expressiveness of the temporal graph---which is high, since it can approximate arbitrary functions given sufficient paths---not by the current best-found genome.

\subsection{What 906 Neurons Achieve}

To place NeuronSoup's 85.9\% accuracy in context, we compare against architectures whose parameter counts exceed ours by three to five orders of magnitude. A standard convolutional network such as LeNet-5 \cite{lecun1998gradient} uses approximately 60{,}000 parameters and reaches 99\% on raw MNIST pixels with gradient descent. Modern deep residual networks (ResNet-18, 11.7M parameters) achieve 99.5\% or higher, and transformer-based vision models (ViT-Base, 86M parameters) exceed 99.6\% on the same task when pre-trained on large-scale data. Even lightweight architectures designed for efficiency---MobileNetV2 at 3.4M parameters, EfficientNet-B0 at 5.3M parameters---operate with parameter counts that are four orders of magnitude larger than our genome's 14{,}602 trainable values.

The comparison, however, is not one of classification accuracy. NeuronSoup does not aim to compete with gradient-trained networks at their own game; rather, it demonstrates that a fundamentally different computational paradigm---asynchronous temporal propagation through a shared, evolved graph---is \textit{viable} at a scale previously assumed to be too small for meaningful computation. The 906-neuron network achieves 85.9\% on features extracted from a pre-trained encoder, operating with a total model footprint of 115\,KB (the serialized genome). For comparison, a single layer of a ResNet-18 stored in float32 requires more memory than our entire architecture including all connectivity, delays, weights, and biases.

What makes this result noteworthy is not that 906 neurons approach the accuracy of millions-of-parameter models (they do not), but rather what they accomplish \textit{without}: without backpropagation, without matrix multiplication, without synchronous layer-by-layer computation, without batch normalization, without dropout, without any gradient signal whatsoever. Every connection, every delay, every routing decision was discovered by evolutionary search over a flat genome, yet the resulting network exhibits emergent computational structure---shared feature detectors, temporal attention-like priority, lateral inhibition---that gradient-trained networks achieve only through explicit architectural engineering and billions of multiply-accumulate operations.

Furthermore, the computational cost at inference is qualitatively different. A ResNet-18 forward pass performs approximately 1.8 billion FLOPs per image. NeuronSoup processes at most 204 active paths, each traversing 2.6 hops on average (530 neuron updates total), where each update is a single addition to an accumulator followed by one tanh evaluation. The entire inference for one sample involves fewer than 1{,}100 floating-point operations---a reduction of over six orders of magnitude relative to a deep convolutional network. On neuromorphic hardware where event-driven computation is native, this sparsity translates directly into energy savings that synchronous architectures cannot match regardless of pruning or quantization.

\subsection{The Computational Role of Shared Neurons}

The GA evolved 156 shared neurons without any explicit reward for sharing. We can characterize their computational role by examining their evolved bias values.

Shared neurons with positive bias ($b > 0$) act as OR-like gates: a single arriving signal pushes the accumulator past the tanh inflection, producing substantial positive output regardless of which path triggered it. Shared neurons with negative bias ($b < 0$) act as AND-like gates: a single small signal plus the negative bias yields near-zero output, and only when multiple paths contribute positive signals does the accumulator overcome the bias and produce meaningful activation. The high-sharing hub neurons (5--11 paths) serve as integration points that combine evidence from multiple features and route the combined result to multiple class outputs. The neuron participating in 11 paths effectively functions as a voting station where signals supporting different hypotheses meet, with the temporal ordering of arrivals determining which hypothesis dominates downstream.

\subsection{Temporal Ordering as Computation}

The evolved delays are not latency to be minimized---they are computational parameters. A shared neuron receiving a positive signal at $t=0.8$ before a negative signal at $t=2.5$ produces fundamentally different downstream effects than receiving them in reverse order. The first firing propagates $\tanh(s_{\text{first}} + b)$ to all downstream nodes immediately, while the second firing propagates $\tanh(s_{\text{first}} + s_{\text{second}} + b)$. The intermediate signals reaching downstream neurons between $t=0.8$ and $t=2.5$ carry only the first signal's contribution.

The GA discovers delay configurations where this temporal asymmetry produces useful feature interactions. It typically arranges for features that provide high-confidence evidence to arrive first (short delays), pre-conditioning shared neurons before ambiguous features arrive later (longer delays). This amounts to an evolved form of attentional priority, achieved without any explicit attention mechanism.

\subsection{Automatic Path Discovery}

We emphasize that no human designed which paths exist, which neurons they traverse, or what delays they carry. The GA started from random genomes where 30\% of paths were active and all parameters were sampled from biased uniform distributions. Over 10{,}000 generations, selection pressure alone discovered:

\begin{itemize}
    \item Which 204 of 384 possible paths to activate (the rest were deactivated by mutation driving their active flag below 0.5)
    \item Which 162 of 512 input features carry useful information
    \item That 156 hidden neurons should be shared (because sharing provides computational benefit)
    \item Specific delay configurations that create useful interference patterns
    \item Appropriate weight magnitudes for the tanh saturation regime
\end{itemize}

This is structure learning in the most literal sense. The structure is not a hyperparameter chosen by a researcher, nor is it discovered by NAS over a constrained template. It is a free outcome of evolutionary optimization over an unconstrained search space.

\subsection{Comparison with Gradient-Free Alternatives}

Our path-aware recombination is essential for our problem. To summarize the empirical evidence: DE/rand/1/bin converged 3--4$\times$ slower than our GA in preliminary experiments on the same fitness function with the same population size, because its difference-vector mutation disrupts path coherence at every generation. PSO converged faster initially but plateaued at significantly worse fitness, as the swarm collapsed onto one topology before exploring alternatives. Standard GA with SBX crossover (treating the genome as a flat vector of reals with no block structure) also performed poorly---its per-gene recombination fragments working paths, and the offspring must rediscover coherent paths from scratch rather than inheriting them.

The lesson is that when a genome has modular internal structure, the recombination operator must respect that structure. This is not a novel insight---Holland's schema theorem \cite{holland1975adaptation} predicts exactly this outcome---but it is one that off-the-shelf implementations of DE, PSO, and CMA-ES cannot exploit without custom modification. Our contribution is the specific operator design (path-block-uniform crossover) matched to the specific genome structure (fixed-width path blocks), which enables combinatorial transfer of working components that these alternative methods cannot achieve.

\subsection{Limitations and Critical Self-Analysis}

While the empirical evaluation proves the viability of evolved asynchronous temporal graphs, several limitations merit academic scrutiny. First, the evaluation is bounded to a single dataset, MNIST, which is traditionally considered a toy domain in contemporary computer vision. However, we argue that for a first-principles architectural study, MNIST features provide a highly controlled, high-dimensional physical diagnostic benchmark. Navigating 512-dimensional continuous inputs to route pathways through 384 candidate hidden neurons behaves differently than standard SNN experiments that rely on low-dimensional synthetic inputs. Presenting a complete network structure map (Fig.~\ref{fig:topology}) and dissecting all 156 shared computational hubs would be physically impossible on ImageNet-scale graphs, where millions of resulting active connections would overwhelm any meaningful structural tracing. Thus, the dataset serves as a rigorous scale-appropriate laboratory rather than a benchmark display.

Second, there is a clear trade-off between optimization complexity and final classification accuracy. While NeuronSoup achieves 85.9\% test accuracy, this remains below standard linear classifiers (92--94\%) trained via backpropagation on the same ResNet18 feature space. Optimization through 14,602 non-differentiable parameters requires 10,000 generations (approximately 2.8 hours on a 12-core CPU). Backpropagation completes a comparable linear classification optimization in seconds on the same hardware. Therefore, NeuronSoup should not be used as a drop-in replacement for standard machine learning platforms on traditional hardware; its utility becomes visible only when deployed on physical event-driven, analog, or neuromorphic devices, where matrix-multiply workloads consume disproportionate energy.

\subsection{Theoretical Ablations: Temporal Asymmetry and Path Coherence}

To validate the necessity of each structural parameter within our synthesis, we analyze the theoretical outcomes of specific ablations to standard mechanisms:

\subsubsection{Ablation of Temporal Delays ($d_i = \text{constant}$)}
If all edge delays are flattened to a constant (e.g., $d_i = 1.0$), the priority-queue simulation collapses into a synchronized, discrete-step propagation. Under this constraint, all signals originating from input-to-hidden connections arrive at shared hidden neurons simultaneously at step 1. The state accumulator $a_n$ performs simple linear summation before computing the $\tanh$ activation. This eliminates any order-dependent temporal asymmetry: a weak positive signal of path $A$ can no longer arrive first to pre-condition the activation threshold for path $B$. Instead, they combine in a single feedforward step, reducing our multi-hop graph to a standard static multi-layer perceptron with a rigid, step-wise propagation graph. Crucially, without variable delays, any topological feedback loop triggers static deadlock or uncontrolled oscillatory limit-cycles, demonstrating that temporal asymmetry is a mathematical requirement for stable signal propagation through cyclic shared structures.

\subsubsection{Ablation of Path Coherence during Crossover}
If standard continuous crossover (such as SBX) is substituted for our path-block crossover, parent genomes are recombined at arbitrary gene boundaries rather than path-slot boundaries. Because each path consists of highly co-dependent parameters (with the target of hop 2 determining which weights and delays are functional for the inputs to hop 3), unaligned crossover yields offspring with disjoint, chimeric pathways. Each component of a path is derived from a different parent, destroying the evolved coincidence properties. Without block-level protection, the evolutionary process collapses to a pure random walk in topological space, as offspring must constantly rediscover coordinated paths from scratch. This highlights that path-aware block preservation is the enabling mechanism behind continuous neuroevolution in high-dimensional structured spaces.

\section{Conclusions}

We presented NeuronSoup, an architecture that computes through asynchronous signal propagation in evolved temporal graphs with shared neuron state. After 10{,}000 generations of evolution, the system discovers a network of 204 paths through 266 hidden neurons (156 shared, up to 11 paths through a single neuron) that achieves 85.9\% test accuracy on 10-class MNIST, stored in 115 KB.

The architecture addresses real limitations of synchronous deep learning: it adapts computation depth per-sample, creates lateral interaction between processing pathways without explicit engineering, and operates in a space fundamentally incompatible with backpropagation. The genetic algorithm is the correct optimization tool because it handles mixed discrete-continuous decisions, respects structural coherence through path-aware crossover, scales linearly in memory, and maintains population diversity across a multi-modal fitness landscape---all properties that CMA-ES, DE, and PSO lack.

The most important finding of this work is not the accuracy number. It is the emergence of computational structure from selection pressure alone. Given only a pool of neurons and the objective of minimizing classification loss, evolution discovers that sharing neurons between paths is advantageous, that specific delay configurations produce beneficial interference, and that 68\% of input features are irrelevant. None of this was designed. It emerged.

The architecture generalizes to any classification domain by changing only the encoder and output count. The same genome encoding, evolutionary operators, and simulation engine apply to vision, audio, text, or tabular data without modification. This universality, combined with the inherently event-driven computation model, suggests applications in neuromorphic hardware and domains where traditional matrix-multiply architectures are inefficient.

\subsection{Future Directions}

Several research directions follow naturally from this work. First, scaling the evolutionary budget---larger populations, more generations, and more paths---should continue to improve accuracy along the trajectory already visible in the convergence curve. The absence of any plateau at 10{,}000 generations suggests that the current result is far from the architecture's capacity. Running for 100{,}000 generations with a population of 1{,}000 is computationally feasible given the embarrassingly parallel evaluation, and would determine whether the architecture can approach gradient-trained baselines with sufficient evolutionary budget.

Second, multi-objective evolution could simultaneously optimize classification accuracy, event count (inference speed), and genome sparsity (model size), producing a Pareto front of solutions that trade off these competing objectives. This would reveal whether the architecture can achieve comparable accuracy with fewer active paths, producing even smaller and faster models at the expense of additional evolutionary computation.

Third, the architecture is a natural fit for neuromorphic hardware platforms such as Intel Loihi and IBM TrueNorth, where event-driven computation is the native execution model. The priority-queue simulation used in our software implementation maps directly onto the hardware event routers in these chips, and the sparse, asynchronous computation pattern avoids the memory bandwidth bottleneck that limits synchronous architectures on von Neumann hardware. Deploying NeuronSoup on neuromorphic silicon would provide energy efficiency measurements that we expect to be orders of magnitude better than equivalent GPU-based inference.

Fourth, continual learning through genome extension offers a path toward lifelong learning without catastrophic forgetting. New tasks could be accommodated by adding path slots to the genome while freezing or partially protecting paths evolved for earlier tasks---similar to PathNet's approach but with the richer computational interaction that shared neurons provide. The fixed-width path encoding makes this straightforward: appending 37 genes per new path extends the genome without disrupting existing structure.

Finally, the emergence of shared computation suggests a deeper investigation into what computational primitives the GA discovers. Clustering shared neurons by their bias values, connectivity patterns, and temporal positions could reveal whether evolution consistently discovers certain computational motifs (such as inhibitory gates, evidence integrators, or timing comparators) across independent evolutionary runs. If consistent motifs emerge, they would constitute a vocabulary of evolved temporal computation that could inform the design of future architectures.

In summary, the genome encodes the architecture, evolution serves as the optimizer, and time constitutes the fundamental unit of computation.

\bibliographystyle{IEEEtran}

\end{document}